\documentclass[11pt]{article}
\usepackage[a4paper,margin=1in]{geometry}
\usepackage{amsmath,amssymb}
\usepackage{booktabs}
\usepackage{graphicx}
\usepackage{float}
\usepackage[numbers,sort&compress,square]{natbib}
\usepackage{hyperref}
\usepackage[table]{xcolor}
\usepackage{microtype}
\usepackage{tabularx}
\usepackage{cji}

\title{Data Scaling as Progressive Coverage of a Predictive Contribution Spectrum}
\cjisubtitle{A suffix-automaton view of predictive spectra and truncation frontiers}
\cjiheadernote{Chunjiang Intelligence}
\author{
\begin{tabular}{cc}
\textbf{Zihui Song} & \textbf{Shihao Ji} \\
\texttt{imbue2025@outlook.com} & \texttt{417818680@qq.com} \\
\\[0.6em]
\textbf{Hongxi Li} & \textbf{Shuaizhi Cheng} \\
\texttt{lihx228@mail2.sysu.edu.cn} & \texttt{szcheng@stu.hit.edu.cn} \\
\\[0.6em]
\multicolumn{2}{c}{\textbf{Chunlin Huang}} \\
\multicolumn{2}{c}{\texttt{7wgn0324@gmail.com}}
\end{tabular}
}
\date{}

\begin{document}
\maketitle

\begin{abstract}
We investigate the hypothesis that real-data scaling laws are governed by progressive coverage of a latent predictive contribution spectrum rather than by token-frequency tails alone. We work with a suffix-automaton representation of text corpora and define a data-intrinsic \emph{global-KL predictive contribution spectrum}, in which each state contributes according to its empirical mass times its KL deviation from a global next-token baseline. Across 12 real corpora, the tail slope of this spectrum is already strongly correlated with the empirical data-scaling exponent of a fixed small GPT learner. We then go beyond slope correlation and define, for each training size $N$, an effective truncation rank $K(N)$ by matching the observed excess loss to the residual tail mass of the prepared 1000k global-KL spectrum. Empirically, $\log K$ is close to linear in $\log N$, with pooled $R^2 \approx 0.96$ for the raw spectrum and $R^2 \approx 0.90$ for the smoothed spectrum. These findings provide strong empirical support for a simple mechanism picture: training scale advances an effective frontier through a predictive state spectrum, and the residual tail mass of that spectrum tracks the remaining excess loss.
\end{abstract}

\section{Introduction}
Scaling laws are often described phenomenologically: loss decreases approximately as a power law in training scale over a broad intermediate regime \cite{kaplan2020scaling,hoffmann2022training}. This description is useful, but it leaves open the central explanatory question: \emph{what object is actually being covered as data scale increases?}

A natural candidate is not token frequency itself, but a deeper predictive state spectrum. In this picture, each dataset induces a ranked family of predictive structures, and training at scale $N$ covers only a finite prefix of that family. The remaining loss is then the residual contribution of the uncovered tail. If this picture is correct, two empirical consequences should follow. First, a state-level predictive spectrum should explain variation in data-scaling slopes better than token-level statistics. Second, actual training scale should correspond to an effective truncation frontier that moves systematically through that spectrum.

This paper reports empirical evidence for both claims. The key object is a suffix-automaton-based \emph{global-KL predictive contribution spectrum}. We show that its tail geometry is strongly tied to cross-dataset scaling behavior, and that empirical loss curves can be reparameterized by an effective spectral cutoff $K(N)$ whose logarithm is approximately linear in $\log N$.

\section{Related Work}
Our empirical picture sits at the intersection of three literatures.

First, the modern empirical study of neural scaling laws established that language-model loss follows regular power-law trends as model size, data size, and compute increase \cite{kaplan2020scaling,hoffmann2022training}. Those works demonstrated the remarkable regularity of the phenomenon, but they remained largely agnostic about the data-intrinsic object whose partial coverage manifests as the observed scaling curve.

Second, a growing theoretical literature has argued that power-law learning curves can often be traced back to spectral structure in the data or in the induced kernel. In particular, Bahri et al.\ connected scaling exponents to spectral decay in resolution-limited regimes \cite{bahri2021explaining}. Our work is aligned with that broader spectral view, but differs in two important respects. We do not work in a teacher--student or kernel-limit setting, and our spectrum is not a covariance or kernel eigenspectrum. Instead, we construct an empirical predictive contribution spectrum directly from text via a corpus-intrinsic state space.

Third, our notion of state is related in spirit to predictive-state viewpoints in dynamical systems, where state is defined through its predictive consequences rather than through a latent hidden variable \cite{littman2001psr}. The suffix automaton gives us a discrete structural analogue of that idea: histories are grouped into end-position classes, and each class can be assigned an empirical next-token distribution. This makes the SAM a particularly convenient bridge between classic string structure and predictive-state reasoning. On the string-algorithm side, suffix automata and directed acyclic word graphs provide a linear-size representation of substring structure and repetition \cite{blumer1985smallest,crochemore2002jewels}. Our contribution is to show that when these states are weighted by empirical mass times predictive KL deviation, they induce a spectrum whose geometry is tightly linked to cross-dataset scaling behavior.

\section{Experimental Setting}
We study data scaling with a fixed decoder-only GPT architecture while varying only the amount of training data. The model uses 6 transformer blocks, 6 attention heads, embedding width 384, context length 128, dropout 0.1, and a GPT-2 vocabulary of size 50,257. Training is performed with AdamW at learning rate $3\times 10^{-4}$ and weight decay 0.1, with batch size 32. For each dataset, we train on five corpus sizes (approximately 100k, 200k, 500k, 1000k, and 2000k tokens) and evaluate by validation cross-entropy. Each run is stopped by an early-stopping criterion based on validation plateauing; the reported loss is the best validation loss achieved along the trajectory.

\begin{figure}[H]
  \centering
  \begin{minipage}{0.82\textwidth}
    \centering
    \includegraphics[width=\linewidth]{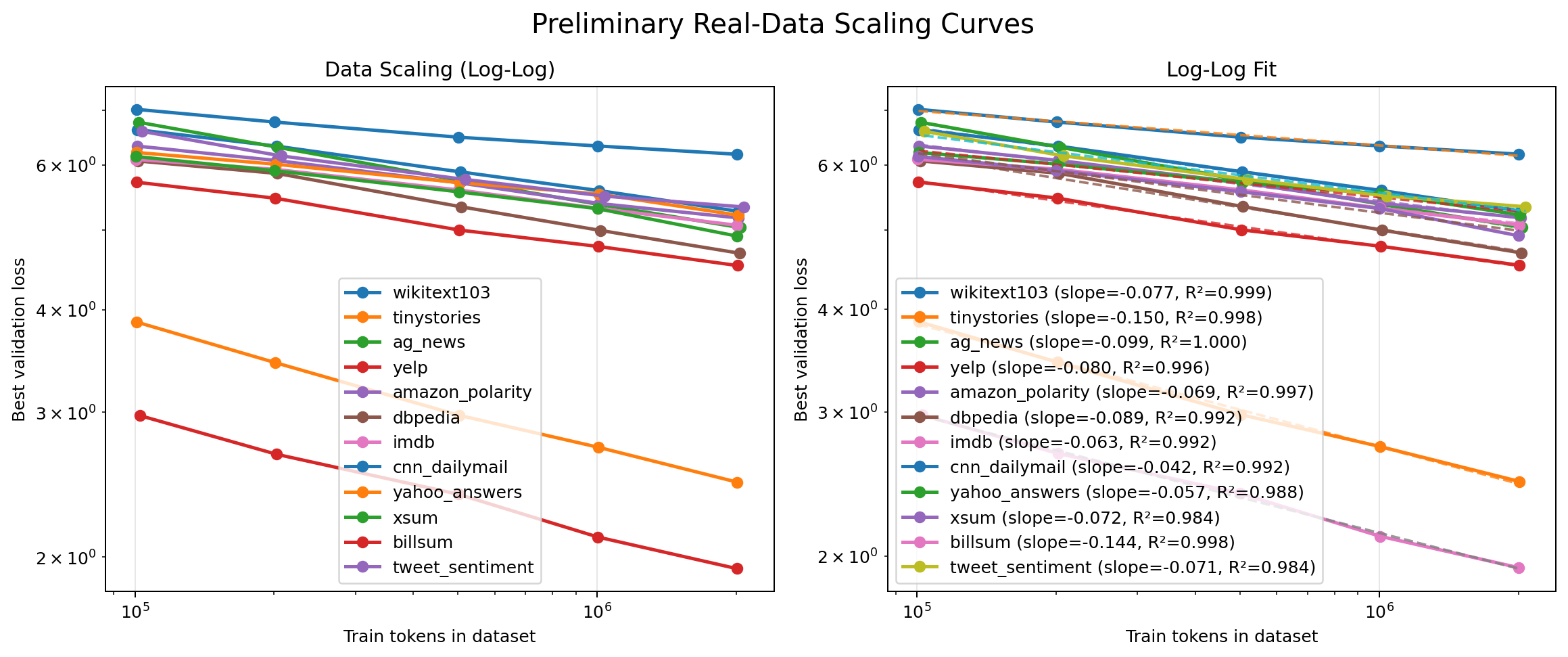}
    \caption{Empirical data-scaling curves for a fixed small GPT learner across real datasets.}
    \label{fig:real-scaling}
  \end{minipage}
\end{figure}

The empirical data-scaling slope for a dataset is then defined as the slope of the least-squares fit of $\log L$ against $\log N$ across these five training sizes, where $L$ denotes best validation loss and $N$ the number of training tokens in the prepared corpus. Representative scaling curves are shown in Figure~\ref{fig:real-scaling}. The benchmark panel includes 12 datasets spanning encyclopedic, narrative, review, classification, and summarization-style domains.

Earlier proxy searches established that simple token-level statistics --- entropy, compression ratio, unigram tail slope, and raw $n$-gram summaries --- are informative but insufficiently robust. Once strongly structured corpora such as TinyStories are included, these scalar summaries no longer provide a convincing mechanism-level account of slope variation. This is the setting in which the predictive spectrum reported below should be interpreted: the aim is not merely to correlate with one dataset family, but to track slope variation across qualitatively different real corpora under a fixed learner.

\section{A Data-Intrinsic State Space}
\subsection{Why the suffix automaton}
\begin{figure}[H]
  \centering
  \begin{minipage}{0.82\textwidth}
    \centering
    \includegraphics[width=\linewidth]{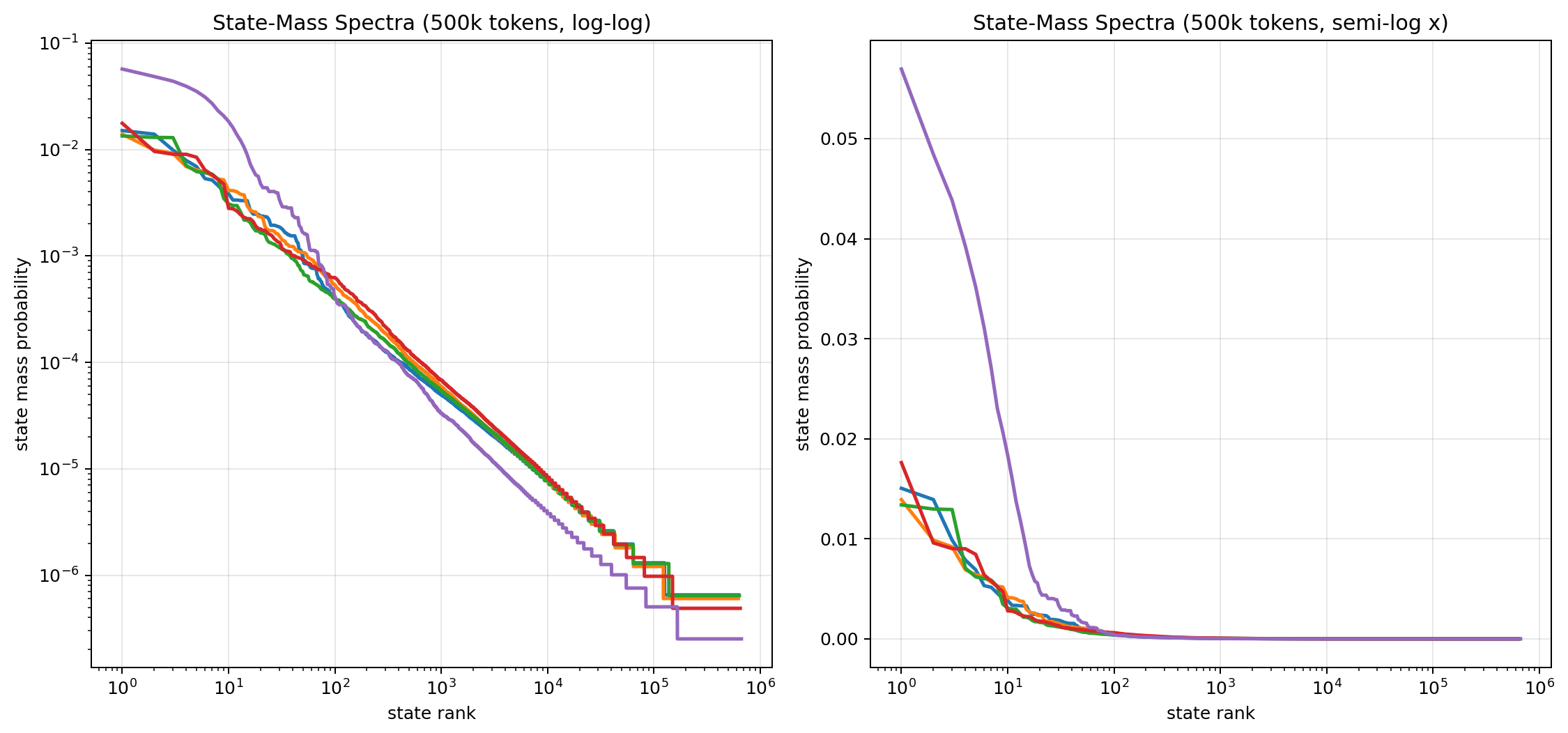}
    \caption{Suffix-automaton state-mass spectra at 500k tokens.}
    \label{fig:state-mass}
  \end{minipage}
\end{figure}

The suffix automaton (SAM) provides a corpus-intrinsic graph representation whose states are end-position equivalence classes of substrings rather than tokens. This makes it a natural structural state space between trivial token counts and fully learned latent representations. The basic state-mass spectrum already exhibits broad power-law structure, as shown in Figure~\ref{fig:state-mass}.

However, state mass alone measures occupancy rather than predictive usefulness. To explain scaling, we need a quantity aligned with reductions in next-token uncertainty.

\subsection{Global-KL predictive contribution}
For each SAM state $s$, define its empirical mass
\[
\mu(s)=\frac{\mathrm{occ}(s)}{\sum_{u \neq 0}\mathrm{occ}(u)},
\]
and let $P(\mathrm{next}\mid s)$ denote the empirical next-token distribution induced by the outgoing transitions of state $s$. Let $P(\mathrm{next})$ be the global empirical next-token distribution, or a smoothed version of it. We define the predictive contribution of state $s$ as
\[
E(s)=\mu(s)\, KL\!\Big(P(\mathrm{next}\mid s)\,\|\,P(\mathrm{next})\Big).
\]
Sorting $E(s)$ in decreasing order yields the \emph{global-KL predictive contribution spectrum}.

This object is attractive for two reasons. First, it remains fully data-intrinsic: no model activations or learned internal states are required. Second, it is naturally aligned with cross-entropy reduction: it measures how much predictive information each structural state contributes beyond the global baseline.

\section{Empirical Spectrum Result}
\begin{figure}[H]
  \centering
  \begin{minipage}{0.98\textwidth}
    \centering
    \includegraphics[width=\linewidth]{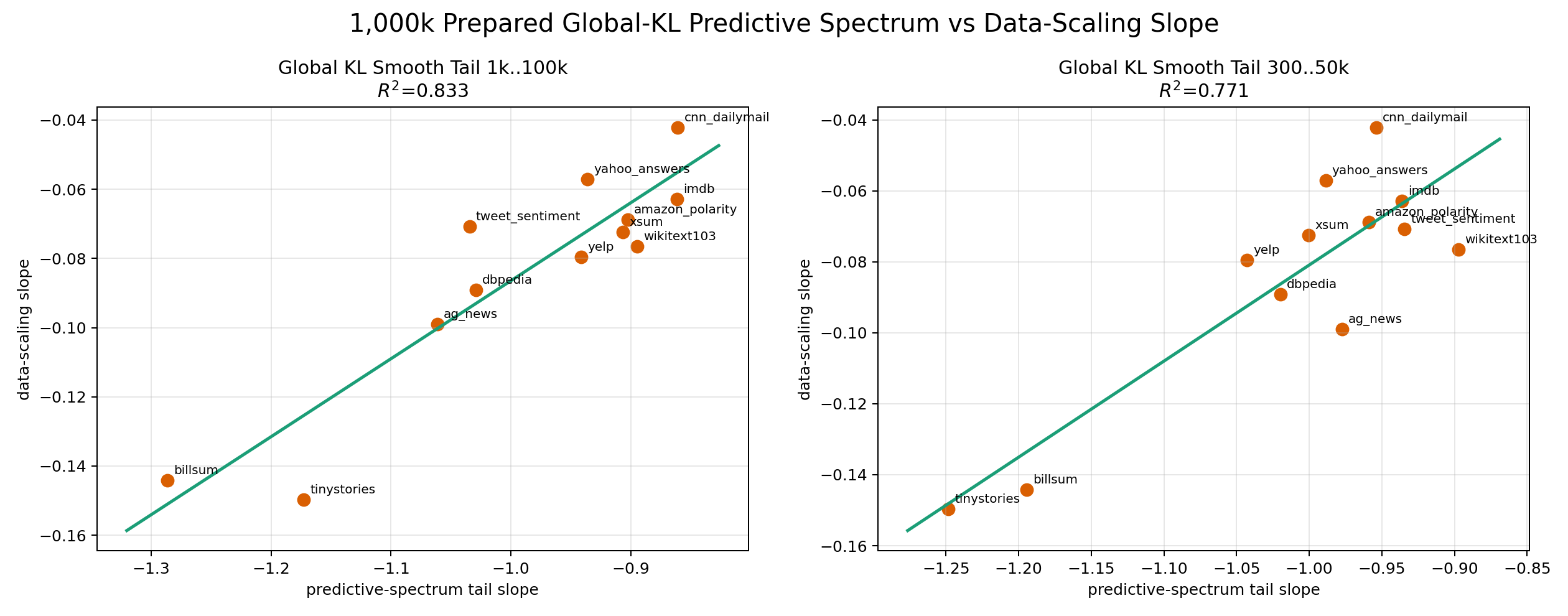}
    \caption{Prepared 1000k global-KL predictive spectrum versus empirical data-scaling slope.}
    \label{fig:gkl}
  \end{minipage}
\end{figure}

Among all proxies tested so far, the prepared 1000k global-KL spectrum is the strongest single-object result. Here ``prepared 1000k'' means that the spectrum is computed once from the dataset's prepared 1000k-token training corpus and then treated as the fixed reference spectrum for that dataset. Figure~\ref{fig:gkl} shows the main comparison against empirical data-scaling slopes.

Among the tail windows emphasized in our prepared-spectrum comparison, the strongest fits are:
\begin{table}[H]
  \centering
  \begin{minipage}{0.62\textwidth}
    \centering
    \caption{Strong tail windows for the prepared 1000k global-KL spectrum.}
    \label{tab:tail-windows}
    \small
    \begin{tabularx}{\linewidth}{>{\raggedright\arraybackslash}Xcc}
      \toprule
      \rowcolor{tableheaderblue}
      Spectrum variant & Fitting window & $R^2$ \\
      \midrule
      Smoothed global-KL tail & $1k \dots 100k$ & $0.833$ \\
      Smoothed global-KL tail & $300 \dots 50k$ & $0.771$ \\
      \bottomrule
    \end{tabularx}
  \end{minipage}
\end{table}

These results are substantially stronger than token-level alternatives and suggest that the relevant explanatory object is a state-level predictive spectrum rather than a vocabulary-frequency tail.

\begin{figure}[H]
  \centering
  \begin{minipage}{0.82\textwidth}
    \centering
    \includegraphics[width=\linewidth]{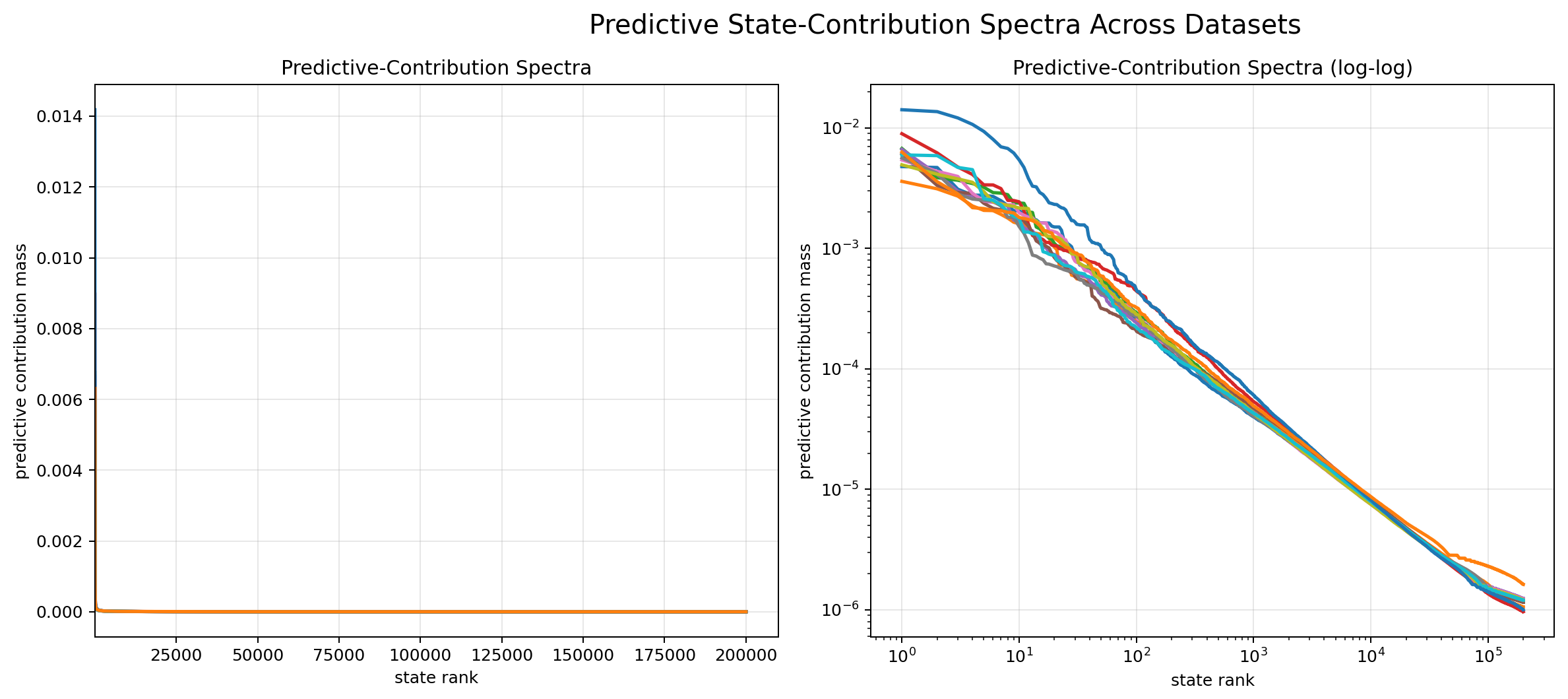}
    \caption{Predictive contribution spectra across datasets.}
    \label{fig:pred-spec}
  \end{minipage}
\end{figure}

The spectra themselves exhibit clear heavy-tailed structure across datasets, as shown in Figure~\ref{fig:pred-spec}.

\section{From Spectrum Slope to Spectral Frontier}
\subsection{Mechanism hypothesis}
The tail-slope result alone remains a cross-sectional correlation. The mechanism claim is stronger: as data scale increases, training should behave as if it progressively covers a prefix of the predictive contribution spectrum. This suggests a decomposition of the form
\[
L(N) \approx \sum_{k > K(N)} w_k,
\]
where $w_k$ denotes predictive contributions sorted in descending order, and $K(N)$ is the effective truncation frontier reached at training size $N$.

Under this view, scaling is not merely a curve fit in $N$; it is a front-propagation process in a ranked predictive state space.

\subsection{Operational definition of $K(N)$}
To make this idea concrete, we define the observed excess loss for each dataset by
\[
\Delta L(N)=L(N)-\min_{N'}L(N').
\]
Let $\Delta L_{\max}=\max_N \Delta L(N)$ for that dataset. Let the prepared 1000k global-KL spectrum be the sorted probability vector $(w_1,w_2,\dots,w_M)$, with $\sum_{k=1}^M w_k=1$, and define the residual tail mass
\[
T(K)=\sum_{k>K} w_k.
\]
We then define the effective cutoff rank by
\[
K(N)=\min\Bigl\{K:\,T(K)\le \frac{\Delta L(N)}{\Delta L_{\max}}\Bigr\},
\]
with the endpoint conventions used in the implementation: the largest observed excess maps to $K=1$, and zero excess maps to $K=M$. In words, the loss at scale $N$ is treated as the amount of predictive contribution mass that remains uncovered beyond rank $K(N)$.

This definition is admittedly operational rather than uniquely principled, but it has the advantage of producing a directly testable prediction: if the frontier picture is correct, then $\log K$ should vary smoothly and perhaps approximately linearly with $\log N$.

\section{Main Frontier Result}
\begin{figure}[H]
  \centering
  \begin{minipage}{0.98\textwidth}
    \centering
    \includegraphics[width=\linewidth]{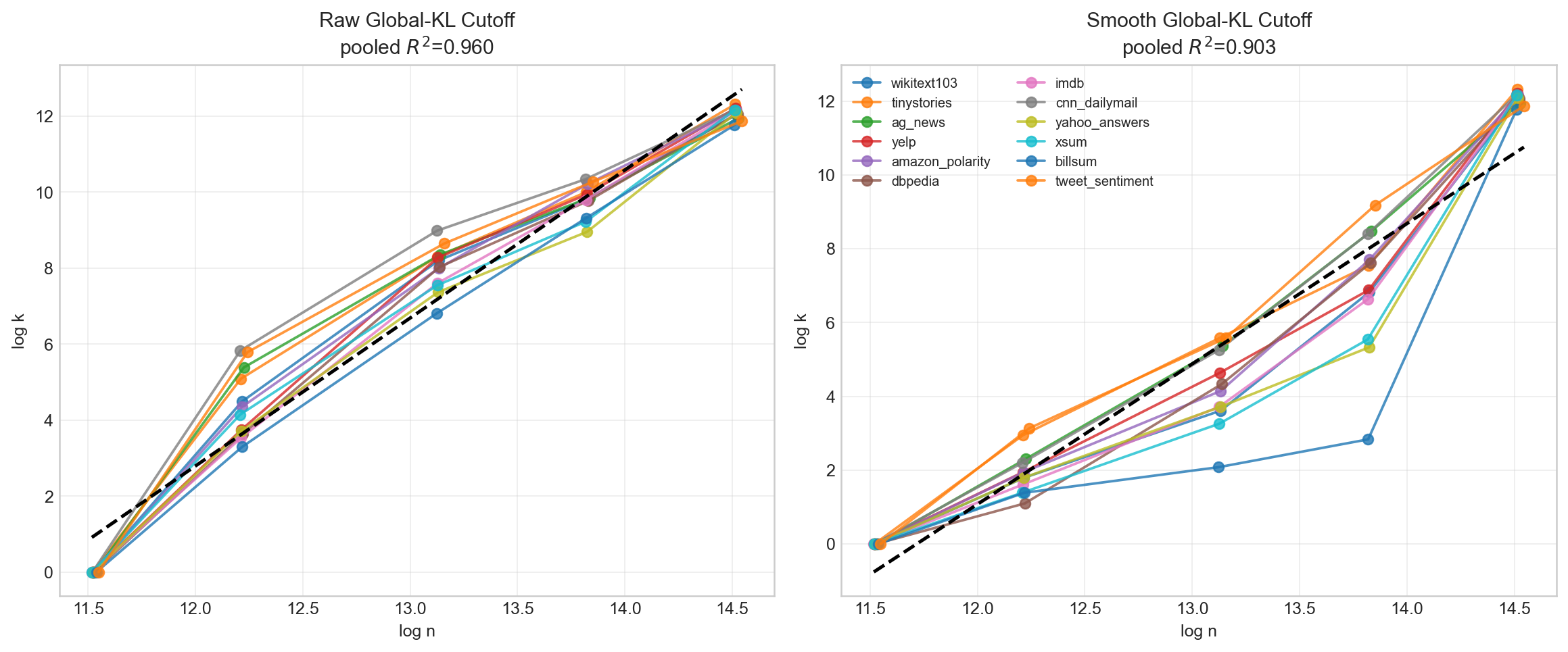}
    \caption{Effective cutoff rank $K(N)$ inferred by matching observed excess loss to residual tail mass in the prepared 1000k global-KL spectrum.}
    \label{fig:trunc-front}
  \end{minipage}
\end{figure}

Figure~\ref{fig:trunc-front} shows the resulting relation between the inferred cutoff rank and training size.

The quantitative summary is reported in Table~\ref{tab:frontier-fits}. The pooled fit uses all 12 datasets and all five training scales per dataset. The raw cutoff construction gives the strongest overall collapse, while the smoothed construction still preserves a high degree of regularity. The lower block of the table reports all dataset-specific smooth fits.

\begin{table}[H]
  \centering
  \begin{minipage}{0.68\textwidth}
    \centering
    \caption{Summary of frontier fits for the inferred cutoff rank $K(N)$.}
    \label{tab:frontier-fits}
    \small
    \begin{tabularx}{\linewidth}{>{\raggedright\arraybackslash}Xcc}
      \toprule
      \rowcolor{tableheaderblue}
      Fit or dataset & Slope & $R^2$ \\
      \midrule
      Pooled raw global-KL cutoff & $3.90$ & $0.960$ \\
      Pooled smooth global-KL cutoff & $3.80$ & $0.903$ \\
      \midrule
      AG News (smooth) & $3.94$ & $0.991$ \\
      Yelp (smooth) & $3.83$ & $0.947$ \\
      Amazon Polarity (smooth) & $3.92$ & $0.955$ \\
      DBPedia (smooth) & $4.01$ & $0.957$ \\
      IMDb (smooth) & $3.81$ & $0.921$ \\
      Tweet Sentiment (smooth) & $3.90$ & $0.992$ \\
      CNN/DailyMail (smooth) & $4.01$ & $0.987$ \\
      Yahoo Answers (smooth) & $3.60$ & $0.869$ \\
      XSum (smooth) & $3.70$ & $0.868$ \\
      BillSum (smooth) & $3.25$ & $0.686$ \\
      TinyStories (smooth) & $3.83$ & $0.970$ \\
      WikiText-103 (smooth) & $3.79$ & $0.924$ \\
      \bottomrule
    \end{tabularx}
  \end{minipage}
\end{table}

This is stronger than a generic proxy correlation. At minimum, it shows that empirical data scaling can be reparameterized by a moving spectral frontier, with training size controlling how deeply the learner penetrates into the ranked predictive state spectrum.

\section{Interpretation}
The empirical picture now admits a simple mechanism chain:
\[
N \mapsto K(N), \qquad
K \mapsto \sum_{k > K} w_k, \qquad
\sum_{k > K} w_k \mapsto L(N).
\]
The first map is empirically close to a power law; the second is the geometry of the predictive contribution tail; the third is the interpretation of excess loss as uncovered predictive mass. Taken together, these results provide evidence consistent with the hypothesis that scaling is governed by progressive coverage of a latent predictive spectrum.

Importantly, this interpretation is more informative than saying merely that ``heavy tails correlate with scaling.'' It suggests a concrete state-space mechanism: the learner appears to advance an effective front through a ranked family of predictive states, while the remaining loss reflects the residual mass of the uncovered tail.

\section{Relation to Quotient-State Refinement}
We also experimented with a more refined state-space construction based on merging SAM states according to similar transition kernels. This quotient construction is conceptually closer to predictive equivalence classes and produces geometrically cleaner spectra:

\begin{figure}[H]
  \centering
  \begin{minipage}{0.82\textwidth}
    \centering
    \includegraphics[width=\linewidth]{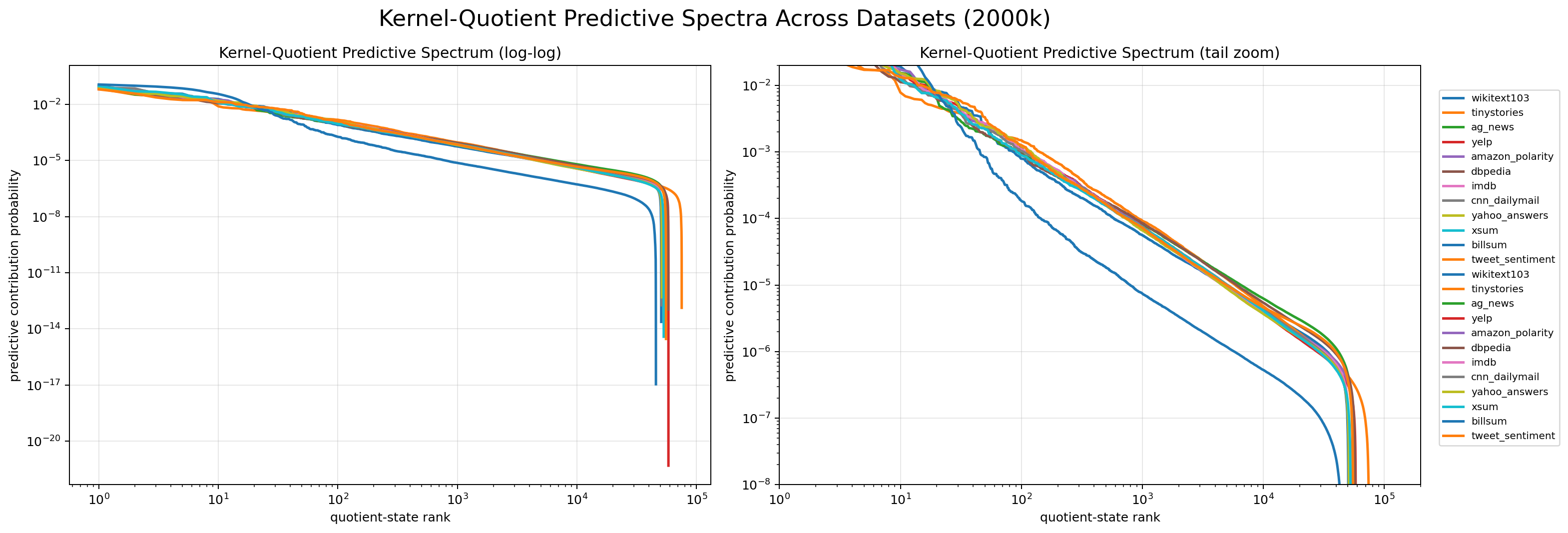}
    \caption{Kernel-quotient predictive spectra at 2000k tokens.}
    \label{fig:kernel-spec}
  \end{minipage}
\end{figure}

However, the explanatory power of the current quotient construction is weaker than that of the unmerged global-KL spectrum. This suggests that the present merging criterion may remove fine-grained information that remains relevant to cross-entropy scaling. In other words, the unmerged spectrum may still be closer to the operative predictive decomposition seen by the learner.

\section{Limitations}
The current frontier result should be interpreted as strong empirical evidence, not as final identification of a unique physical quantity.

First, the present cutoff rule has visible endpoint anchoring: the smallest-$N$ point often maps to $K \approx 1$, while the largest-$N$ point often saturates near the available tail length. This means that part of the apparent linearity is constrained by construction. What remains encouraging is that the interior regime is still highly structured.

Second, the current definition of excess loss is normalized independently within each dataset. This makes cross-dataset comparison cleaner, but also leaves open the question of whether alternative normalizations or alternative loss floors would produce the same frontier relation.

Third, the predictive contribution spectrum is still an operational proxy for a deeper latent object. The strongest current evidence is that it is a particularly good proxy; it does not yet establish that it is the uniquely correct ontological notion of a learnable mode.

\section{Conclusion}
The main empirical conclusion is simple. A suffix-automaton-based global-KL predictive contribution spectrum not only explains variation in data-scaling slopes across datasets; it also supports a much stronger picture in which the effect of data scale is to move an effective truncation frontier through that spectrum. Empirically, the inferred frontier obeys an approximately linear relation between $\log K$ and $\log N$ over a broad range. This does not by itself prove that the global-KL spectrum is the unique latent object behind data scaling, but it does provide unusually direct evidence that scaling behavior can be interpreted as progressive coverage of a ranked predictive state spectrum.

\begingroup
\small
\setlength{\itemsep}{0pt}
\setlength{\parskip}{2pt}

\endgroup

\end{document}